\documentclass[runningheads]{llncs}
\usepackage{graphicx}
\usepackage{algorithm,algorithmic}
\usepackage{stackengine}
\usepackage{gensymb}
\begin{document}
\title{Automatic segmentation of skin lesions using deep learning}
\titlerunning{Automatic segmentation of skin lesions}
%
\author{Joshua Peter Ebenezer\inst{1}\and
Jagath C. Rajapakse\inst{2}}
\authorrunning{Ebenezer and Rajapakse}
%
\institute{Indian Institute of Technology, Kharagpur, India
\\ \and
Nanyang Technological University,  Singapore}
\maketitle              
\begin{abstract}
This paper summarizes the method used in our submission to Task 1 of the International Skin Imaging Collaboration's (ISIC) Skin Lesion Analysis Towards Melanoma Detection challenge held in 2018. We used a fully automated method to accurately segment lesion boundaries from dermoscopic images. A U-net deep learning network is trained on publicly available data from ISIC. We introduce the use of intensity, color, and texture enhancement operations as pre-processing steps and  morphological operations and contour identification as post-processing steps. 

\end{abstract}
\section{Introduction}
The International Skin Imaging Collaboration\footnote{https://isic-archive.com} (ISIC), an international effort to improve melanoma diagnosis, holds a recurring challenge every year for the developoment of image analysis tools for segmentation of dermoscopic images. This paper summarises our submission to Task 1 of the challenge, which is lesion boundary segmentation. The data released for the ISIC 2018 challenge were used for our submission~\cite{ref_article1,ref_article2}. No private datasets or other publicly available datasets were used.
\par
In our submission, we used a deep learning architecture and an image processing pipeline for  accurate segmentation of skin cancer lesions from dermoscopic images. Images of lesions generally have poor contrast and boundaries of the lesions are often hard to outline as lesion color is similar to the color of the skin near the edges. The pre-processing is able to significantly enhance the information in the images and suppress noise.

\section{Methodology}

\subsection{Intensity contrast enhancement}
Skin lesion images are color images. The contrast enhancement algorithm based on the layered-difference representation of 2D histograms proposed by Lee at al.~\cite{ref_article3} is used to enhance the contrast of the intensity of the image, where intensity is defined as the arithmetic mean of the red (R), blue (G), and green (G) channels. 

\subsection{Hue-preserving color enhancement}
Color provides details of skin lesions and following Naik and Murthy~\cite{ref_article4}, we apply a hue-preserving color enhancement technique to enhance the contrast of the color images. 
Hue is the attribute of color according to which an area of an image appears similar to a perceived color. Color enhancement techniques generally involve non-linear transformations between color spaces, which lead to the gamut problem when pixel values go out of the acceptable bounds of the RGB space during conversion, which causes undesirable changes in the hue. The resulting dermoscopic images have enhanced color, their hue is preserved, and they have clearer lesions.

\subsection{Texture enhancement}

We use the multiscale texture enhancement algorithm based on fractional differential masks proposed by Pu et al.~\cite{ref_article5}. The second mask proposed is used as it has the best precision, convergence, and texture enhancement properties, according to the paper. 
8 convolution kernels, one for each direction, are created for rotation invariance as suggested by Pu et al., with fractional order $v=0.5$, and applied to the intensity of the images. The magnitude of the response is found using the Euclidean norm. 

\subsection{U-net architecture} 
\begin{figure}
\centering
\includegraphics[scale=0.4]{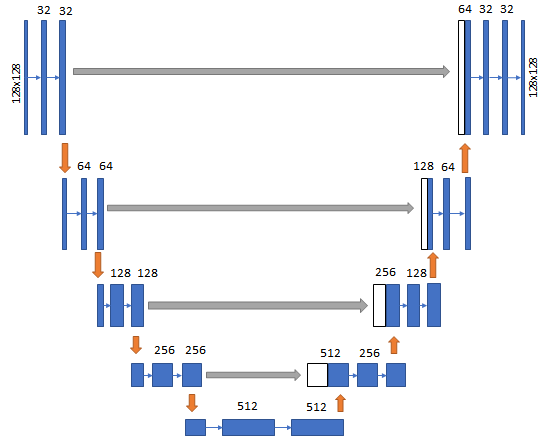}
\caption{Illustration of the modified U-net architecture.} \label{fig1}
\end{figure}

A modified architecture of U-net~\cite{ref_article6} is used for this study. The primary difference from the original architecture is that the output segmentation map produced is of the same size as the input image, unlike in the original architecture, and the number of filters is half of that in the original architecture at each stage. The modified architecture is shown in Fig.~\ref{fig1} and consists of a contracting path on the left and an expanding path on the right. All input images are resized to 128x128 using bilinear interpolation. The output of the network are sigmoid probabilities on a 128x128 map. After the network is trained, the output is converted to a binary labeled image at a threshold of 0.5 applied to logit probabilities.

\subsection{Post-processing operations} 
The sigmoid probability map generated by the network does not take into account the connectivity of segmented masks. Consequently, there are holes and small blobs in the segmented image. Opening (erosion followed by dilation) and closing (dilation followed by erosion) with a 5x5 square structuring element are used to remove these. The largest connected contour in the image is found and drawn as the predicted mask. After post-processing, the image is resized to its original shape.

\section{Experiments and results}
The dataset consists of 2594 dermoscopic images and corresponding masks and the images are of different sizes. 34 different random splits were performed on the data released, and 30 U-net networks, each with different random initializations as well, were trained on these splits. 3 randomly selected groups of 25 networks each were used, whose outputs were averaged to create three sets of submissions.

\subsection{Pre-processing and augmentation}
Aggressive on-line data augmentation was performed before contrast enhancement. The images were allowed a random rotation range, flipping, and zooming. The proposed method is to train the network with five channels: three color enhanced channels, a contrast-enhanced intensity channel, and a texture-enhanced intensity channel.  

\subsection{U-net training}
The Adam optimizer~\cite{ref_article7} was used for training the U-net with a learning rate of 1e-4, regularization parameters $\beta_1=0.9$ and $\beta_2=0.999$, and no decay in the learning rate over each update. The loss was defined as the negative of the Dice similarity coefficient\footnote{For given sets $X$ and $Y$,  Dice coefficient = $\frac{2|X \cap Y|}{|X|+|Y|}$}. The Jaccard index\footnote{Jaccard index = $\frac{|X \cap Y|}{|X \cup Y|}$} was monitored as a metric. A mini-batch size of 4 was used and the training was run for 200 epochs. Models were saved at checkpoints of best Jaccard coefficient on random validation splits for each network. The Keras 2 library\footnote{https://keras.io} with a Tensorflow backend was used for the training and evaluation and all experiments were run on a 16 GB Nvidia Tesla P100 GPU. 

\subsection{Results on validation data}
The three submissions received thresholded Jaccard indices of 0.756, 0.750, and 0.746 on the validation set of 100 images. The thresholded index is defined as 0 if the Jaccard index if less than 0.65, and equal to the Jaccard index if more than 0.65.

\section{Conclusion}
We presented a deep learning image processing framework that achieved state-of-the-art results. We used five channel inputs to train the U-net architecture: the contrast-enhanced intensity channel, the three hue-preserved color-enhanced RGB channels, and the texture-enhanced intensity channel;  the U-net architecture was modified to make both input and output image sizes same; and a post-processing method that uses morphological operations and contour identification was used to arrive at the output segmentation. The framework is fast, simple, and efficient and can be extended to other applications as well.

\subsection*{Acknowledgement}
This work is partially supported by AcRF Tier-1 grant RG149/17 by the Ministry of Education, Singapore

\end{document}